\documentclass[runningheads]{llncs}
\usepackage[T1]{fontenc}
\usepackage{xcolor}
\usepackage{hyperref}
\usepackage{algorithm} 
\usepackage{algpseudocode} 
\usepackage{cite}
\usepackage{amsmath,amssymb,amsfonts}
\usepackage{graphicx}
\usepackage{textcomp}
\usepackage{xcolor}
\usepackage{hyperref}
\usepackage{fancyhdr}
\usepackage{graphicx}
\usepackage{url}
\usepackage{graphicx}
\usepackage{amsmath}
\usepackage{amssymb}
\usepackage{xcolor}
\usepackage{enumitem}
\usepackage{lipsum}
\usepackage{multirow}
\usepackage{array}
\usepackage{makecell}
\usepackage{diagbox}

\usepackage{pifont}

\newcommand{\bharath}[1]{\textcolor{black}{#1}}

\algnewcommand\algorithmicinput{\textbf{Input:}}
\algnewcommand\Input{\item[\algorithmicinput]}

\algnewcommand\algorithmicoutput{\textbf{Output:}}
\algnewcommand\Output{\item[\algorithmicoutput]}

\newlist{inlinelist}{enumerate*}{1}
\setlist[inlinelist]{label=(\arabic*)}
\newlist{inlinelistsub}{enumerate*}{1}
\setlist[inlinelistsub]{label=(\roman*)}

\begin{document}

\title{ReFit: A Framework for Refinement of Weakly Supervised Semantic Segmentation using Object Border Fitting for Medical Images}
%
%
\author{Bharath Srinivas Prabakaran\inst{1}\thanks{These authors contributed to this work equally.}
\and
Erik Ostrowski\inst{1}$^\star$
\and
Muhammad Shafique\inst{2}}

%
\authorrunning{ }
%
\institute{Institute of Computer Engineering, Technische Universit{\"a}t Wien (TU Wien), Austria\\
\email{\{bharath.prabakaran,erik.ostrowski\}@tuwien.ac.at}\\
\and
eBrain Lab, Division of Engineering, New York University Abu Dhabi (NYUAD), United Arab Emirates (UAE)\\
\email{muhammad.shafique@nyu.edu}}

\maketitle              
\fancypagestyle{firstpage}
{
    \fancyhead[C]{To appear at the 18th International Symposium on Visual Computing (ISVC’23), October 2023, Lake Tahoe, NV, USA.}    
    \fancyhead[R]{}
}
\thispagestyle{firstpage}

\begin{abstract}
Weakly Supervised Semantic Segmentation (WSSS) relying only on image-level supervision is a promising approach to deal with the need for Segmentation networks, especially for generating a large number of pixel-wise masks in a given dataset.
However, most state-of-the-art image-level WSSS techniques lack an understanding of the geometric features embedded in the images since the network cannot derive any object boundary information from just image-level labels. 
We define a boundary here as the line separating an object and its background, or two different objects.
To address this drawback, we are proposing our novel \textit{ReFit} framework, which deploys state-of-the-art class activation maps combined with various post-processing techniques in order to achieve fine-grained higher-accuracy segmentation masks. 
To achieve this, we investigate a state-of-the-art unsupervised segmentation network that can be used to construct a boundary map, which enables \textit{ReFit} to predict object locations with sharper boundaries.
By applying our method to WSSS predictions, we achieved up to 10\% improvement over the current state-of-the-art WSSS methods for medical imaging. 
The framework is open-source, to ensure that our results are reproducible, and accessible online at \url{https://github.com/bharathprabakaran/ReFit}.
\end{abstract}

\begin{keywords}


Weak Supervision, Semantic Segmentation, Masks, Medical Imaging,

Framework, CAM, Activation Maps, Boundary, Refinement
\end{keywords}



\section{Introduction}


State-of-the-art approaches for the semantic segmentation task require a deep learning model trained on a large amount of heavily annotated data, a task referred to as Fully Supervised Semantic Segmentation (FSSS). The generation of pixel-wise masks for large datasets is a resource- and time-consuming process. 
For instance, one frame of the Cityscapes dataset, which contains thousands of pixel-wise frame annotations of street scenes from cities, requires more than an hour of manual user-driven annotation~\cite{CT}.
Furthermore, medical imaging and molecular biology fields require the knowledge of highly qualified and experienced individuals capable of interpreting and annotating the images. 
The interpretation of these individuals may vary based on their experience and know-how, further inducing a problem of varying masks, known as annotator bias.
Therefore, to reduce the time and resources required for generating pixel-wise masks, a wide range of research works focus on developing approaches that focus on weaker kinds of supervision. 
This is where Weakly Supervised Semantic Segmentation (WSSS) can be highly beneficial.

WSSS approaches focus on generating masks with minimum supervision, such as image-level labels~\cite{IL1,IL2,IL3}, bounding boxes~\cite{BB1,BB2}, point annotations \cite{POINT}, and scribbles~\cite{SC1, SC2}.
This work focuses on generating semantic segmentation masks for medical images using the most straightforward, and least-supervised, image-level labels.
\bharath{We limit the scope of our work to just using image-level labels since they are the most inexpensive form of annotation.}
To the best of our knowledge, the current state-of-the-art in image-level-based WSSS methods use class activation maps (CAMs)~\cite{CAM} to generate the pixel-level masks of an object from its image-level label.
The central idea of CAMs is to use any model trained with classification loss to generate activation maps that highlight the image regions responsible for the prediction decision.
This results mostly in a rough localization of the objects rather than precise pixel-wise masks. 
The most popular CAM approaches focus on adding regularization loss to improve the quality of the CAM prediction~\cite{SUB, PUZZLE} or utilizing refinement methods that aim to enhance the CAM afterward~\cite{AFF, SEAM}.
For example, adversarial erasing~\cite{ADV1} erases the most discriminative part of the CAM to force the model to consider different parts of an object.
Chang et al.~\cite{SUB} use clustering to automatically sub-divide every class into sub-classes, implicitly generating distinctive classes for less discriminative parts of the CAM.
\bharath{Several other techniques focus on exploiting the ability of state-of-the-art classifiers to obtain viable object masks for the dataset~\cite{xu2022multi,zhou2022regional,ru2022learning}.
However, these approaches are not primarily feasible for medical imaging use-cases, due to the classifier's inability to learn relevant information from the smaller and complex datasets.}
We propose to investigate the applicability of CAMs for automating the generation of pixel-wise segmentation masks for medical images given their image labels.

\begin{figure*}[t]
\centering
\includegraphics[width=0.8\textwidth]{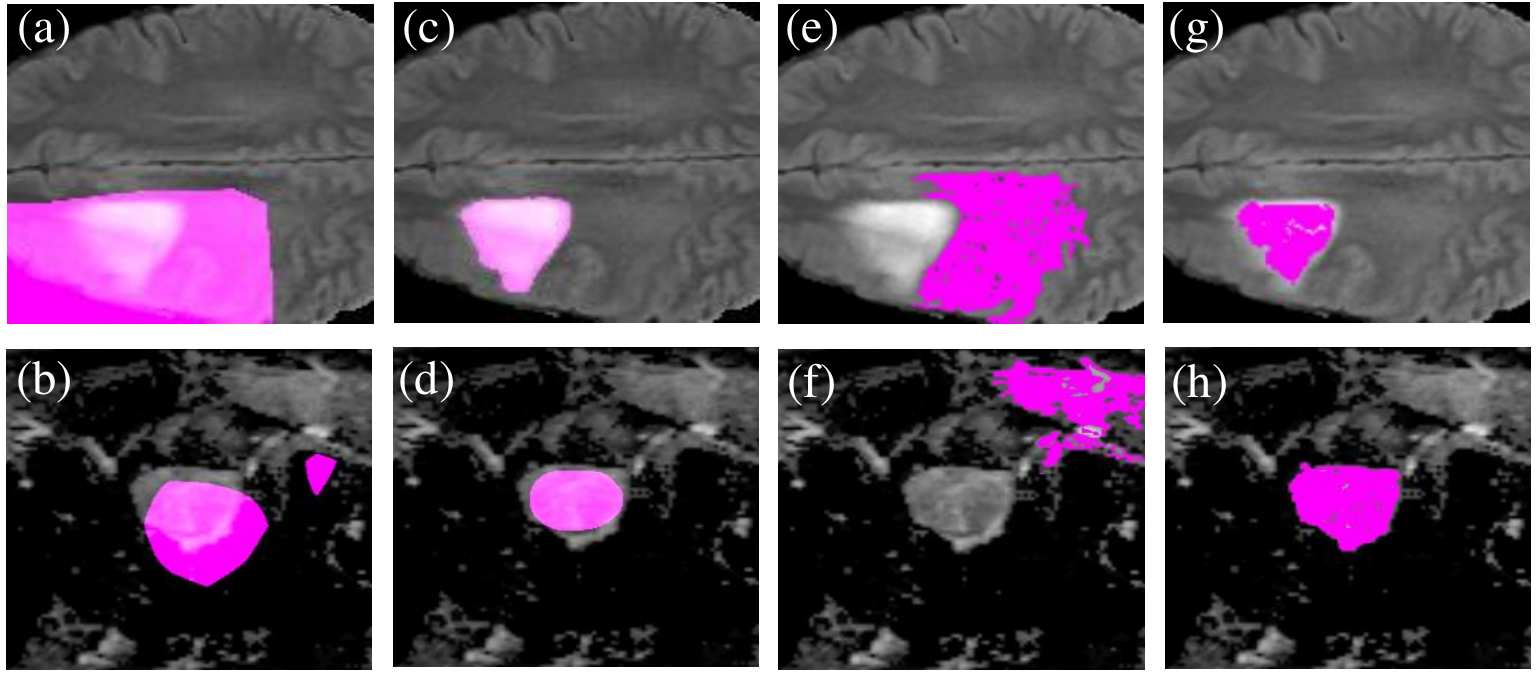} 
\caption{Comparing the CAM's masks (a and b) and  their ground truths (c and d) against the use of similarity-based clustering strategies when incorrect (e and f) and accurate (g and h) anchors are identified on the BRATS and Decathlon datasets.\label{fig1}}
\end{figure*}

Although recognizing only the most discriminative parts of an object is enough to achieve a high classification accuracy, the activation map might leave out a sizeable portion of the object or include irrelevant background information, thus leading to an inaccurate mask. 
The most common approaches to acquiring a complete segmentation mask include adding more demanding conditions to the loss function, which improves the output CAM. 
This can be further refined, with respect to the initial CAM output, usually with pixel similarity calculations.
Although more complex loss functions may succeed in covering the complete object, their downside is that the highlighted area remains blurry and is still far from the sharp predictions achieved using FSSS. 
On the other hand, the refinement-based methods can predict sharp object borders, but they heavily rely on the quality of the initial predictions since they cannot spot parts of the object with a different color gradient than the rest. 
This could also lead to potential scenarios wherein a generated mask is inaccurate in certain instances when incorrect anchors are used to generate the masks with respect to an identified object.
Fig.~\ref{fig1} presents an overview of these differences with the help of illustrative examples from the BRATS~\cite{menze2014multimodalBRATS1} and Decathlon~\cite{antonelli2022medicalDecathlon} datasets. 
As shown, the CAM prediction successfully generates a mask almost around the complete object but includes a lot of background pixels. 
In contrast, a similarity-based pixel clustering instance achieves sharper borders but may fail to capture the object due to changes in color gradients with respect to the background.
It might also fail entirely in cases where wrong anchors are used due to a bad base prediction.

Therefore, state-of-the-art frameworks combine the two approaches to help alleviate each other's shortcomings to generate segmentation labels, known as \textit{pseudo-labels}, to train an FSSS network, which proved to be the most accurate.
For medical imaging datasets, state-of-the-art CAM models are not effective because the classifier fails to learn on them due to the smaller size and complexity of the dataset.
Hence, we improve upon this strategy by extracting information obtained from saliency maps and their boundaries, to differentiate between the object and its surroundings through our novel BoundaryFit module that acts as an additional step between the initial CAM prediction and mask refinement.


As we have already discussed, the improved CAM output generally captures the complete object but still includes many background pixels in its predictions. 
These background pixels tend to deteriorate the quality of the refinement method, if they achieve higher logits/certainty than the actual target object.
This may occur if the classifier fails to make confident predictions on the given dataset.
Hence, our proposed BoundaryFit module aims to refine the borders of the initial CAM prediction without relying on pixel similarities and instead focuses on Unsupervised Semantic Segmentation (USS) and saliency methods to generate a simplified \textit{edge map} of the input image, which is the first step of our framework.
Second, we use the edge map to determine the image's \textit{response map} using GradCAM \cite{GRAD}, which is a network used to obtain CAMs. 
Finally, we combine the two using our BoundaryFit module, which refines the object boundaries to obtain a fine-grained segmentation mask.
We perform extensive experiments on three different medical imaging datasets, namely, the breast cancer ultrasound (BUSI)~\cite{BUSI}, the BRATS 2020~\cite{menze2014multimodalBRATS1,bakas2018identifyingBRATS2}, and the Decathlon~\cite{antonelli2022medicalDecathlon} datasets, to prove the effectiveness of the proposed framework and illustrate the benefits of our approach.
To sum up, the key contributions of this work are:
\begin{enumerate}[leftmargin=*]
    \item Our novel ReFit framework can generate fine-grained segmentation masks of medical images using just image-level labels. 
   \item The BoundaryFit module, which can be incorporated into any conventional WSSS pipeline to obtain fine-grained segmentation masks. 
   \item  We have illustrated the benefits and improvements of using ReFit on three real-world medical imaging datasets. 
   The ReFit framework, including the BoundaryFit module, is open-source and accessible online\footnote{\url{https://github.com/bharathprabakaran/ReFit}}.
\end{enumerate}

\begin{figure*}[t]
\centering
\includegraphics[width=\textwidth]{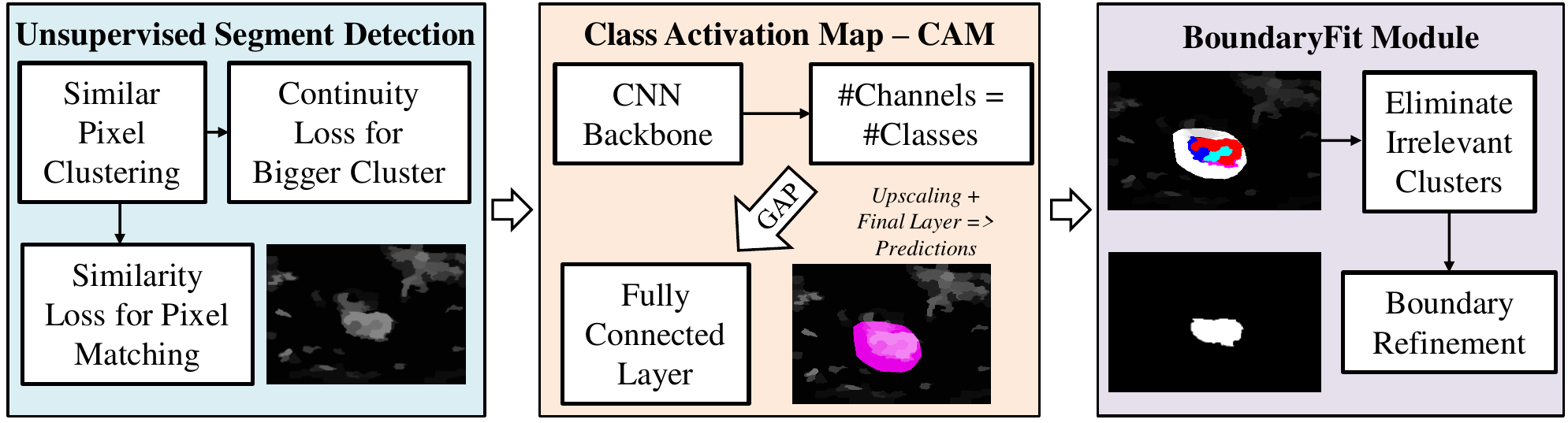} 
\caption{Overview of our ReFit Framework, which starts with the unsupervised segmentation (USS) stage clustering the pixels of an image based on their similarity to the surrounding pixels; Next, we extract the edge map based on the input cluster, which determines a preliminary boundary of the target object; Finally, our BoundaryFit module conforms the two to obtain a fine-grained mask. \label{fig2}}
\end{figure*}

\section{Our ReFit Framework}

Fig.~\ref{fig2} presents an overview of our framework, which currently incorporates a generic CAM to obtain the necessary activation maps. 
First, an unsupervised segmentation (USS) network is used to cluster the image pixels into bigger groups based on the Quickshift \cite{QUICK} and SLIC~\cite{SLIC} algorithms to create superpixels.
Second, we use a trained classifier to generate a CAM, which acts as a preliminary segmentation mask, based on clusters from the USS, for each class in a given input image.
Followed by which our BoundaryFit module fits the fuzzy CAM prediction to the boundaries outlined by the USS map to generate a refined segmentation mask for each object in the image. 
If necessary, these pseudo-labels can be subsequently used to train a fully supervised semantic segmentation network to improve its precision.

\subsection{Unsupervised Segment Detection}

Our approach focuses on pre-refining the CAM prediction to the nearest edge inside the mask in order to reduce the model's identification of false-positive pixel predictions.
To achieve this, we propose to generate a helpful edge map using a USS network to break down the image into simpler parts as a function for the input $I$ as $I_{EM}=g_{USS}(I)$.
The USS will cluster similar pixels together, removing unimportant details in the input image, leaving us with a less detailed image. 
The USS method we use for image simplification consists of two convolutional layers with ReLU, Batch Norm, a continuity loss, and a similarity loss. 
The continuity loss prevents the network from using an arbitrary number of pixel clusters and constrains them with a user-defined maximum value $q$.
Meanwhile, the similarity loss encourages the network to only cluster pixels close to each other in the feature space.

For the ReFit framework, we used SLIC and Quickshift as the continuity loss and performed an extended hyperparameter search to generate suitable simplified images. 
On the one hand, clustering too many pixels into one group will merge the object with the background, while clustering too few pixels together will not simplify the image enough.
Note that optimizing the hyper-parameters for a particular image would not work for a different image. 
Therefore our goal was to find hyperparameters that work well enough for all images. 

To elaborate further, we iteratively cycle through all possible hyperparameter values in small step-sizes, while evaluating their results on the dataset to obtain the parameter value that offers the best results for a small sample of up to $100$ random images.
we have observed such a sample represents the dataset very well.
The hyperparameter search could be mathematically optimized to find the optimal combination if an annotated dataset for validation is provided. 
But sufficient hyperparameters can be found by testing a few images and evaluating their performance. 
We can find better hyperparameters when using an annotated dataset for validation. 
But sufficient hyperparameters can be found by testing less than $100$ images and evaluating their performance. 
The goal is to simplify the images as much as possible without losing the shape of the target objects.
The use of other clustering algorithms is orthogonal to our approach and can easily be included in the framework.


\subsection{Class Activation Map - CAM}
Next, we start by training a conventional classifier model on the target dataset, which is subsequently used by a GradCAM~\cite{GRAD} to extract the response map for each input edge map obtained from the USS stage. 
The generated segmentation masks for each class $n$ in $I_{EM}$ are considered together to be the initial response map, which is defined as $M_{CAM}:=\{M_{CAM}^0,M_{CAM}^1, ...,M_{CAM}^n\}$.
The original CAM predictions highlight the most discriminative object parts and a lot of background pixels around them, as illustrated in Fig.~\ref{fig1}.
For example, the initial response map from our CAM covers a significant portion of the image to ensure that the complete object is inside along with some background pixels. 
The next step, \textit{i.e.,} the BoundaryFit module, is used to eliminate the irrelevant pixels and obtain a fine-grained segmentation mask of the object.

\subsection{The BoundaryFit Module}

Once we successfully generate the edge map, we can use it as an additional supervision source, which provides us with geometrical guidance for improved boundary detection. 
Hence, we can combine it with the initial response map from CAMs, which usually captures the complete object at the cost of adding many background pixels.
Next, with a combination of the edge map obtained from the previous stage and a Floodfill algorithm~\cite{FLOOD}, we successfully remove some irrelevant background pixels from the mask.

The Floodfill algorithm accomplishes this by starting from any negative pixel $(x,y)$ in the initial response map and then turning all pixels belonging to the same cluster as $(x,y)$ negative as well.
Note that we define a cluster as the set of pixels inside an area closed by the boundaries in an edge map $I_{EM}$, including the pixels that demarcate the image borders. 
Since some pixel clusters also reach into the CAM prediction, only pixel clusters completely inside the CAM mask will remain positive. We define this step as follows:

{
\begin{equation}
\begin{split}
h_{BF}(M_{CAM}, I_{EM})&=(M_{CAM}(I_{EM}) \otimes I_{EM})\\
&=M_{BF}
\end{split}
\end{equation}}

\noindent where $h_{BF}(\cdot)$ and $\otimes$ denote our BoundaryFit module and the combination process of edge map and CAM prediction, respectively, which provides us with the final segmentation mask ($M_{BF}$) that fits the estimated boundaries of the object.
\section{Results and Discussion}


\bharath{The experiments are completed on a CentOS 7.9 Operating System executing on an Intel Core i7-8700 CPU with 16GB RAM and 2 Nvidia GeForce GTX 1080 Ti GPUs.
We executed our scripts using CUDA 11.5, Pytorch 3.7.4.3, torchvision 0.11.1, and Pytorch-lightning 1.5.1.
We use the Dice Similarity Coefficient (DSC) metric, as shown by the state-of-the-art~\cite{PATEL2022102374CMER, SEAM}, to illustrate the benefits of using our approach to generate segmentation masks for the medical imaging datasets. 
We also consider the mean Intersection-over-Union (mIoU) ratio as an evaluation metric to analyze the framework's benefits.}
For all experiments, we evaluate the Dice Similarity Coefficient as:

\begin{equation}
    \centering
    DSC = \frac{2 | A \cdot B|} {|A|+|B|}
\end{equation}

\noindent where $A$ and $B$ are binary matrices denoting the ground truths and classification results, respectively, with each element corresponding to $1$ for elements inside a group and $0$ otherwise.
We also evaluate the mean Intersection-over-Union (mIoU) ratio, which is used as the evaluation metric, and defined as

{
\begin{equation}
\centering
mIoU = \frac{1}{N} \sum_{i=1}^N \frac{ p_{i,i}}{\sum_{j=1}^N p_{i,j} + \sum_{j=1}^N p_{j,i} - p_{i,i}  }
\end{equation}
}

\noindent where $N$ is the total number of classes, $p_{i,i}$ the number of pixels classified as class $i$ when labelled as class $i$. $p_{i,j}$ and $p_{j,i}$ are the number of pixels classified as class $i$ that were labelled as class $j$ and vice-versa, respectively.

We have evaluated our framework on the following three different medical imaging datasets:
\begin{inlinelist}
\item the breast cancer ultrasound dataset BUSI~\cite{BUSI},
\item the Multimodal Brain Tumor Segmentation dataset BRATS 2020~\cite{menze2014multimodalBRATS1,bakas2018identifyingBRATS2,bakas2017advancingBRATS3}, and
\item the Decathlon dataset~\cite{antonelli2022medicalDecathlon}.
\end{inlinelist}
For Decathlon, the dataset is divided in a $3:1$ ratio for generating the training and validation sets, respectively; the subdivided dataset is already accessible as part of the code repository published in the WSS-CMER work. 
Every image contains either just the background or the background and the singular positive class. 
The base classifiers were trained on the training set and then evaluated on the validation set. 
The decathlon dataset contained the T2 and ADC phases for each image. 
We only used the ADC phase as discussed in the paper.
For the BraTS dataset, the 3D images of the dataset were divided into slices and cropped. 
All slices were randomly subdivided into one of three sets to perform three-fold cross-validation. 
For training, the classifiers used a combination of two sets for training and were validated on the remaining set.
For the BUSI dataset, since we do not train a model on the data, we directly evaluate the ImageNet pre-trained model on it. 
BUSI contains three classes, normal, benign, and malignant. 
We excluded the malignant class for evaluation since our method struggled to detect the malignant cases, as stated in the main article.

\begin{table}[t]
\centering 
{

\begin{tabular}{lcc} 
\hline
Method                       & Avg. DCS & Avg. mIoU\\ \hline
Blank                        & 64.1 & 47.2\\
GradCAM                      & 65.5 & 48.5\\
ReFit                  & \textbf{67.2} & \textbf{50.6}\\ \hline
\end{tabular}}
\caption{Comparison of GradCAM and ReFit (w/~GradCAM backbone) on BUSI.\label{tab1}}
\end{table}

\begin{figure}[t]
\centering
\includegraphics[width=0.5\textwidth]{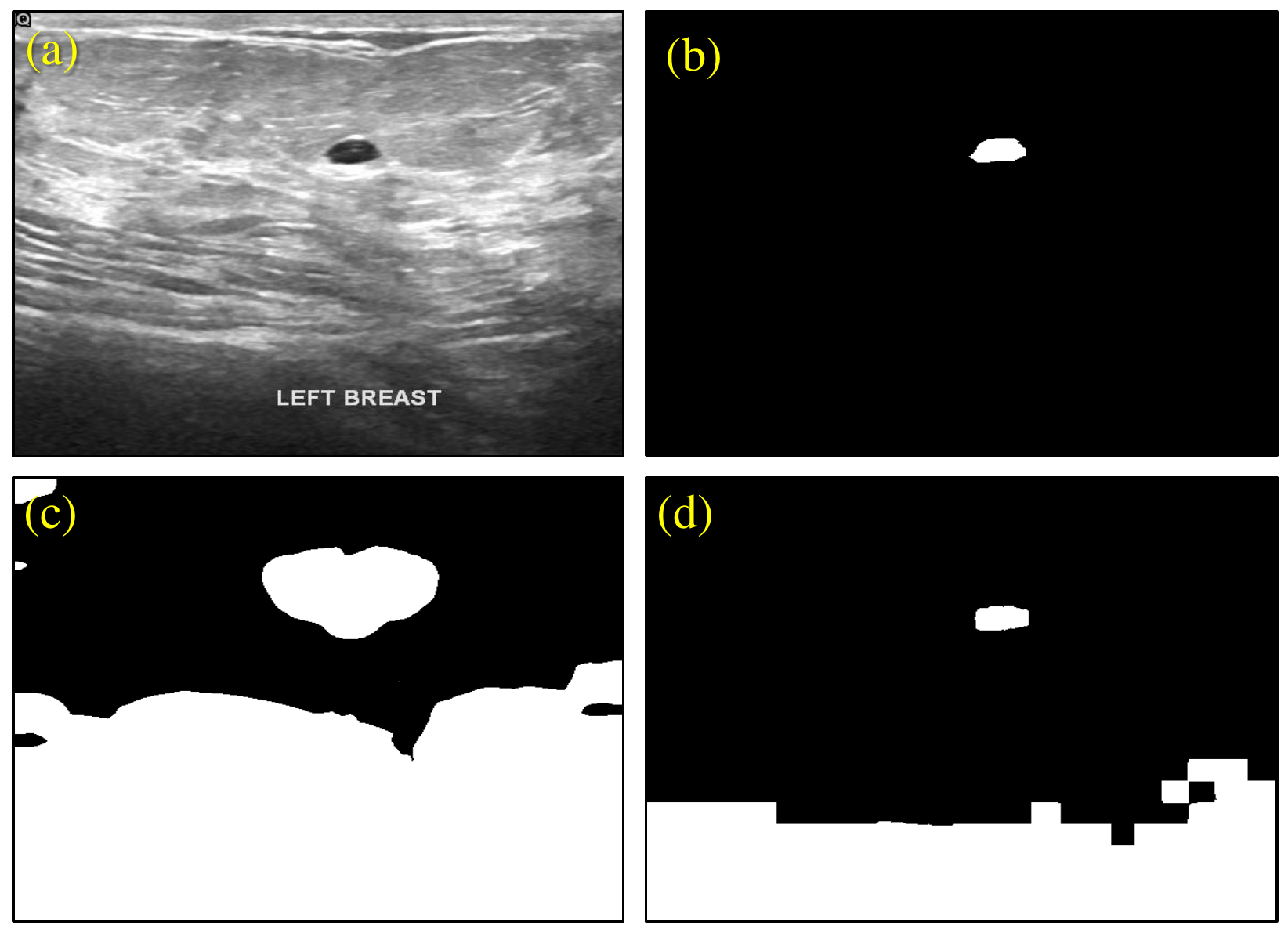}
\caption{(a) Image from the breast ultrasound dataset~\cite{BUSI}; (b) ground truth; (c) GradCAM prediction; (d) ReFit prediction.\label{fig4}}
\end{figure}

\textbf{BUSI Dataset:} We start by examining the results of GradCAM, built using a state-of-the-art classifier for the dataset, and compare them with the results from ReFit.
Note, since learning the other classes proved too difficult with the low amount of data, we have only evaluated the benign cases.
We use an ImageNet~\cite{IMAGENET} pre-trained GradCAM with a ResNet34~\cite{RESNET} backbone as a baseline, instead of a state-of-the-art CAM model for natural images, because the GradCAM achieved better results on the dataset. 
This is probably because the BUSI dataset is relatively small and does not contain enough data for a larger model to learn all the relevant information effectively.
After generating the response maps with GradCAM, we have refined them using our BoundaryFit module to achieve the results presented in Table~\ref{tab1}.

Conventional user-level evaluation of the BUSI images provides us with the understanding that the cancer objects in each image are relatively small compared to the background (see Fig.~\ref{fig4}).
Therefore, predicting a simple blank mask achieves an mIoU of $47.2$\%, while using GradCAM improves the results to $48.5$\%. 
However, with the ReFit framework, we improved the best prediction mask to the highest possible value. 
Similarly, the average DCS for ReFit is the highest among all three with a value of $67.2\%$. 
The state-of-the-art SEAM~\cite{SEAM} and WSS-CMER~\cite{PATEL2022102374CMER} techniques are unable to learn sufficient information from this dataset to make predictions. 

\begin{table}[t]
\centering 
{\fontsize{9pt}{9pt}\selectfont
\begin{tabular}{lcc}
\hline
Method         & Avg. DCS & Avg. mIoU \\ \hline
SEAM~\cite{SEAM}                  & 56.1 & 39.0\\
WSS-CMER~\cite{PATEL2022102374CMER}              & 59.7 & 42.6\\
\bharath{Blank} & 64.6 & 47.7\\
GradCAM (w/ ResNet34)     & 67.3 & 50.7\\
GradCAM (w/ ResNet50)     & 68.5 & 52.1\\
ReFit (w/ ResNet34)   & 70.3 &  54.2\\ 
ReFit (w/ ResNet50)   & \textbf{71.2} &  \textbf{55.7}\\ \hline
FSSS & 81.8 & 69.2 \\\hline
\end{tabular}}
\caption{Comparison of ReFit with state-of-the-art techniques on BRATS.\label{tab2}}
\end{table}

\begin{table}[t]
\centering 
{\fontsize{9pt}{9pt}\selectfont
\begin{tabular}{lcc}
\hline
Method         &  Avg. DCS & Avg. mIoU\\ \hline
\bharath{Blank} & 64.6 & 47.7\\
GradCAM (w/ ResNet101)    & 65.9 & 49.1 \\
SEAM~\cite{SEAM}                  & 65.9 & 49.1 \\
GradCAM (w/ ResNet18)     & 67.0 & 50.3\\
WSS-CMER~\cite{PATEL2022102374CMER}              & 71.3 & 55.4\\
GradCAM (w/ ResNet50)     & 76.1 & 61.4\\
GradCAM (w/ ResNet34)     & 78.0 & 63.9\\
ReFit (w/ ResNet34)      & \textbf{79.4} & \textbf{65.8} \\ \hline
FSSS & 86.8 & 76.7 \\\hline
\end{tabular}}
\caption{Comparing ReFit and state-of-the-art techniques on Decathlon.\label{tab3}}
\end{table}


\begin{figure*}[t]
\centering
\includegraphics[width=0.8\textwidth]{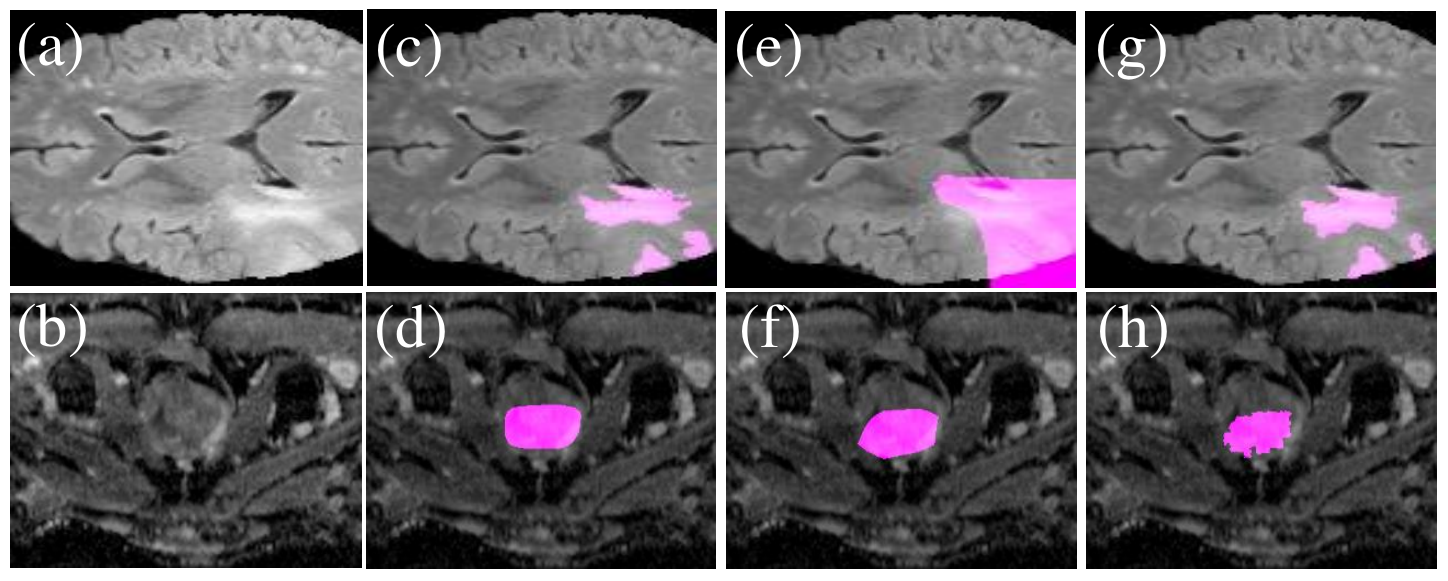}
\caption{Visual comparison of the BRATS and Decathlon dataset images (a and b, respectively); their ground truth segmentation masks (c and d); GradCAM predictions (e and f); and ReFit predictions (g and h).\label{fig:Ab}}
\end{figure*}

\textbf{BRATS 2020 Dataset:} Next, we evaluate the effectiveness of our ReFit framework on the BRATS 2020 dataset and compare it with state-of-the-art techniques like SEAM and WSS-CMER.
These results are shown in Table~\ref{tab2}.
Like with the BUSI dataset, we additionally evaluate the effectiveness of a GradCAM constructed using trained ResNet34 and ResNet50 classifiers as the baseline. 
As expected, our ReFit framework (with ResNet50) outperforms all other techniques to achieve average DCS and mIoU values of $71.2\%$ and $55.7\%$, respectively.
\bharath{
The FSSS evaluation is the result reached when using a traditional, fully supervised semantic segmentation network is trained on the provided ground-truth pixel-wise masks.
This network acts as the upper bound, when compared to the use of other WSSS approach using pseudo-labels,, i.e., the goal of WSSS approaches is to reach or surpass the accuracy of FSSS.
}
Like in the WSS-CMER~\cite{PATEL2022102374CMER} approach, we have combined all instances into one single `positive' class.
We have followed a similar approach for the Decathlon dataset, which is the next use-case.

\textbf{Decathlon Dataset:} Finally, we analyze the efficacy of the our framework on the Decathlon Prostate dataset and compare it with the state-of-the-art and simple GradCAMs built using various classifier variants. 
Out of all these approaches, the best outcome was achieved using a GradCAM built with a ResNet34 model, surpassing the state-of-the-art approaches.
Hence, we evaluate the ReFit framework with a ResNet34-based GradCAM, which achieves the best outcome with a DCS score of $79.4\%$.
We have illustrated the results of these experiments in Table~\ref{tab3}.

\bharath{
Although we have illustrated that ReFit is successful in improving upon the current state-of-the-art, the literature on WSSS for medical imaging is not comprehensive and comparison partners are rather lacking. 
We are working with a set of very low-quality state-of-the-art predictions, trying to improve them to a certain degree. 
More investigation might be required into this to determine the actual benefits of using CAMs and the BoundaryFit module. 
Similarly, the BoundaryFit module works on the understanding that the object is clearly distinguishable from the background based on its boundaries; in unlikely scenarios where this is no longer the case, this could lead to potential problems.
}

Note, the state-of-the-art results for the last two datasets were extracted directly from \cite{PATEL2022102374CMER}, since we were unable to reproduce them at our end.

\subsection{Ablation Studies}
\label{sec:Ablation}

Fig.~\ref{fig:Ab} illustrates the visual comparison of the ReFit predictions for the BRATS and Decathlon datasets.
As anticipated, the GradCAM over-estimates the object mask by including a significant number of background pixels for both datasets (Figs.~\ref{fig:Ab} (e) and (f)).
And, by design, a significant portion of these pixels are eliminated by our BoundaryFit module to obtain a fine-grained segmentation mask (Figs.~\ref{fig:Ab} (g) and (h)).


\section{Conclusion}

In this paper, we have proposed the novel ReFit framework, which introduces a novel BoundaryFit module between CAM and the final FSSS model, to achieve finer segmentation masks, which can be used to improve the overall accuracy of the model. 
The BoundaryFit module provides additional saliency by utilizing unsupervised semantic segmentation models, which refine the CAM predictions to obtain higher-quality training labels for state-of-the-art FSSS models.
The BoundaryFit module can be incorporated into any pre-existing WSSS framework to boost the quality of its predictions, as illustrated by three real-world medical imaging datasets. 
Finally, we showed that the predictions generated by ReFit achieve state-of-the-art performance, proving its effectiveness compared to other approaches by up to $10\%$.
ReFit is open-source and is accessible online at \url{https://github.com/bharathprabakaran/ReFit}.










\section*{Acknowledgments}
This work is part of the Moore4Medical project funded by the ECSEL Joint Undertaking under grant number H2020-ECSEL-2019-IA-876190.
This work was also supported in parts by the NYUAD’s Research Enhancement Fund (REF) Award on “eDLAuto: An Automated Framework for Energy-Efficient Embedded Deep Learning in Autonomous Systems”, and by the NYUAD Center for Artificial Intelligence and Robotics (CAIR), funded by Tamkeen under the NYUAD Research Institute Award CG010.


\bibliographystyle{ieeetr}
\bibliography{emenblebib}
\end{document}